\documentclass[10pt,twocolumn,letterpaper]{article}

\usepackage{wacv}              

\usepackage{multirow}
\usepackage[table]{xcolor}
\usepackage[accsupp]{axessibility}
\usepackage{tabularx}

\definecolor{wacvblue}{rgb}{0.21,0.49,0.74}
\usepackage[pagebackref,breaklinks,colorlinks,allcolors=wacvblue]
{hyperref}

\def\wacvPaperID{1774} 
\def\confName{WACV}
\def\confYear{2027}

\title{TERA: Task-Evidence Restoration for Underwater Object Detection under Bandwidth-Constrained Transmission}

\author{
Heejin Kim\\
Sungkyunkwan University\\
Seoul, Republic of Korea\\
{\tt\small hijin1998@skku.edu}
\and
Joon-Seok Song\\
Sungkyunkwan University\\
Seoul, Republic of Korea\\
{\tt\small jundol@skku.edu}
\and
Eunil Park\thanks{Corresponding author}\\
Sungkyunkwan University\\
Seoul, Republic of Korea\\
Jaume I University\\
Castellon, Spain\\
{\tt\small eunilpark@skku.edu}
}

\begin{document}
\maketitle

\begingroup
\renewcommand{\thefootnote}{}
\footnotetext{%
  \copyright{} 2027 IEEE. Personal use of this material is permitted. Permission from IEEE must be obtained for all other uses, in any current or future media, including reprinting/republishing this material for advertising or promotional purposes, creating new collective works, for resale or redistribution to servers or lists, or reuse of any copyrighted component of this work in other works.
}
\endgroup

\begin{abstract}

Underwater object detectors are typically evaluated on pre-transmission images, overlooking additional degradation from bandwidth-constrained downscaling and compression.
We introduce AquaTX, a controlled benchmark built from real underwater detection datasets with seen and held-out codecs.
We also propose TERA, a metadata-blind receiver-side framework combining task-guided pixel restoration and dual-state feature correction while keeping the detector frozen.
Under seen compound degradation, TERA improves mean average precision (mAP$_{50:95}$, averaged over IoU thresholds from 0.50 to 0.95) from 37.7 to 50.0 on DUO-TX and from 31.5 to 41.3 on RUOD-TX relative to the frozen detector without restoration.
TERA also preserves clean-input performance and remains competitive with detector fine-tuning under the held-out AVIF codec.
These findings support transmission-aware evaluation and task-guided restoration for underwater object detection.
Code is available at \url{https://github.com/hijinkim/TERA}.
\end{abstract}    
\section{Introduction}
\label{sec:intro}

Underwater robots are increasingly used for marine ecosystem monitoring, fishery resource management, and infrastructure inspection, where object detection is essential for automatically analyzing collected imagery~\cite{er2023research,liu2021new,ditria2020automating,xu2023systematic,jian2024underwater}.
In bandwidth-constrained underwater monitoring systems, spatial downscaling and lossy compression can be applied to reduce the amount of transmitted image data~\cite{akyildiz2005underwater,yuan2021low,li2023rfd}.
Consequently, detectors encounter transmission-induced degradation in addition to the optical degradation inherent to underwater imaging~\cite{zhu2026toward,wu2025underwater}.
We focus on degradation introduced by resizing and lossy coding rather than channel-level corruption such as packet loss or bit errors.
Spatial downscaling reduces the spatial support of small objects, while lossy compression can suppress local structures relevant to detection. 
As transmission settings vary, these effects can alter detection outcomes even for the same underlying scene~\cite{gandor2022first,zhang2024deep,dodge2016understanding,poyser2021impact}. 
This motivates a controlled evaluation that isolates transmission-induced performance loss from pre-existing underwater degradation.

\begin{figure}
    \centering
    \includegraphics[width=1\linewidth]{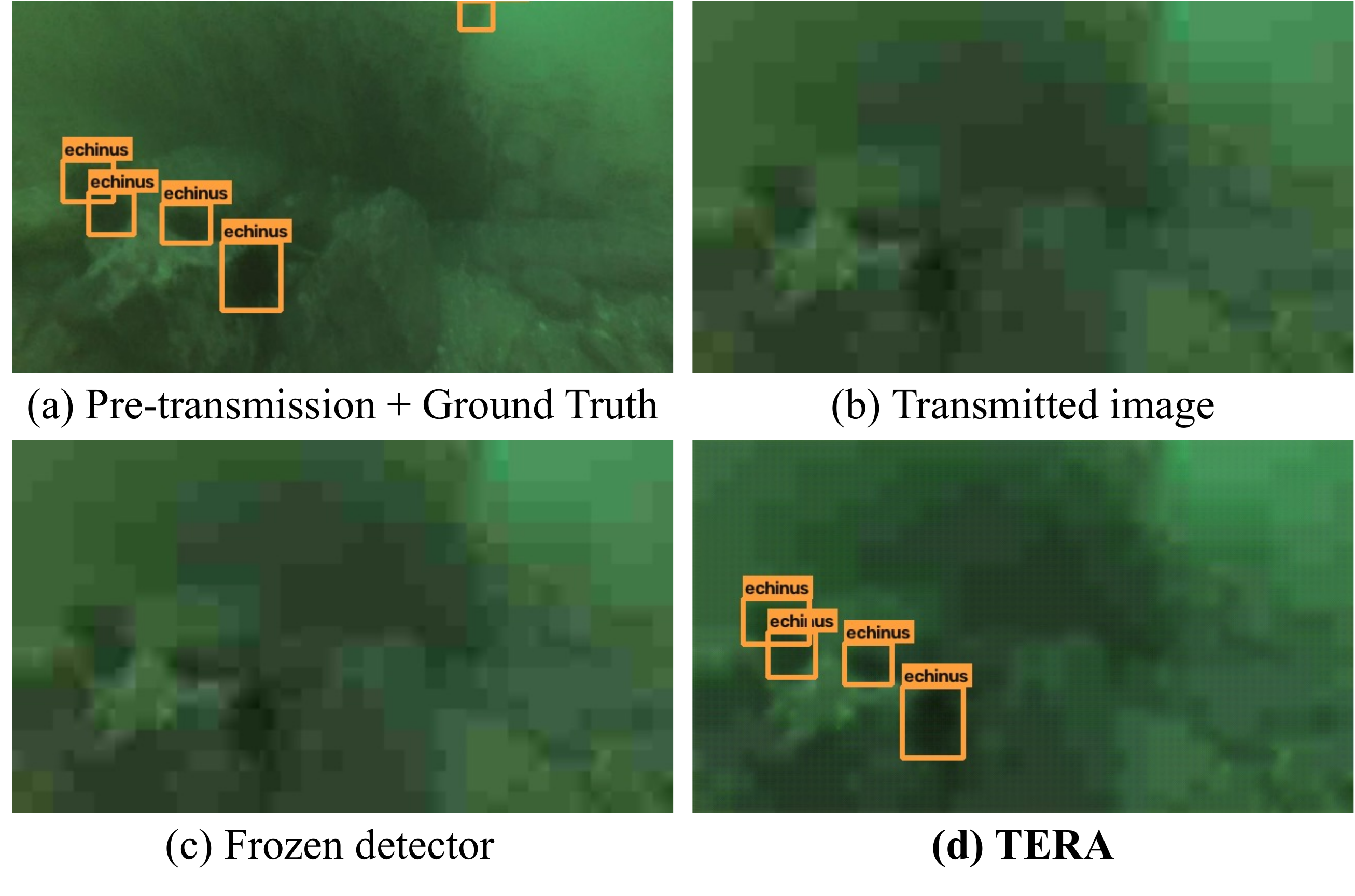}
    \caption{
    Detection under bandwidth-constrained transmission.
    (a) Pre-transmission image with ground truth.
    (b) Transmitted image after downscaling and compression.
    (c) Missed detections by the frozen detector.
    (d) TERA restores task evidence and recovers the missed detections without updating the detector.
    }
    \label{fig:motivation}
\end{figure}

Unlike conventional underwater enhancement, receiver-side restoration targets task-relevant information lost during transmission rather than remove the underwater appearance already present in the detector's training data.

Existing Underwater Image Enhancement (UIE) methods primarily target optical degradation~\cite{zheng2026comprehensive}, while some compression-aware restoration methods assume known degradation parameters~\cite{li2026hpgn}.
Underwater low-bit-rate methods instead modify or jointly optimize the codec, requiring control over both transmitter and receiver~\cite{wu2025underwater}.
These approaches do not directly address restoration under varying standard codecs with a fixed detector.
Moreover, existing underwater detection benchmarks lack controlled transmission degradations for isolating their impact on detection performance.

We address these gaps with \textbf{AquaTX} and \textbf{TERA}.
AquaTX applies controlled resize--compression operators to real underwater detection datasets and provides fixed, hash-verified assets for reproducible evaluation across seen and unseen codecs.
TERA restores transmission-damaged task evidence through image and detector-feature restoration while keeping the detector frozen and requiring no transmission metadata at inference.

The main contributions of this work are:
\begin{itemize}
    \item We introduce \textbf{AquaTX}, a benchmark that isolates transmission-induced degradation in real underwater detection data through controlled resizing and standard-codec compression, with unseen-codec evaluation and fixed assets for reproducible comparison.

    \item We propose \textbf{TERA}, a metadata-blind framework that restores transmission-damaged task evidence in both image and detector-feature spaces while keeping the detector fully frozen.
    Its dual-state feature restoration jointly exploits degraded and pixel-restored detector features to recover residual task evidence.

    \item We demonstrate on DUO-TX and RUOD-TX that TERA recovers substantial detection performance under compound transmission while preserving clean-input performance and remaining competitive under an unseen codec.
\end{itemize}
\section{Related Work}
\label{sec:related_work}

\subsection{Underwater Image Enhancement and Task-driven Restoration}
Underwater image enhancement (UIE) mitigates optical degradation such as color distortion, contrast loss, and blur~\cite{anwar2020diving}.
While conventional UIE focuses primarily on perceptual quality~\cite{akkaynak2019sea,li2019underwater,li2021underwater,peng2023u}, recent task-driven methods optimize enhancement for downstream recognition.
DTI-UIE~\cite{lin2026downstream} introduces task-aware attention and perceptual supervision for detection and segmentation, TACL~\cite{liu2022twin} uses task-aware feedback, and detection-perceptor~\cite{chen2020perceptual} approaches propagate recognition supervision to the enhancer.
Related task-driven restoration has also been explored for adverse imaging conditions~\cite{kim2025exploiting,liu2022image} and recognition-oriented super-resolution~\cite{haris2021task}.

These methods demonstrate the benefit of optimizing restoration for machine perception, but do not specifically address receiver-side degradation caused by bandwidth-constrained underwater transmission.
In contrast, TERA targets information lost through spatial reduction and lossy compression during transmission, without adapting the detector or requiring transmission metadata.

\subsection{Feature-space Restoration and Compression-aware Perception}
Feature-space approaches denoise corrupted activations~\cite{xie2019feature,yang2022self} or transfer clean representations to networks operating on degraded inputs, while some compression-aware restoration methods exploit codec information such as quality factors or quantization parameters~\cite{li2026hpgn}.
Underwater low-bit-rate methods jointly design learned codecs to preserve perceptual or machine-vision quality~\cite{wu2025underwater}.

These approaches either adapt the recognition model, rely on known degradation information, or require control over the transmission codec.
TERA instead operates entirely at the receiver: it keeps the detector frozen, uses only decoded RGB inputs, and combines image restoration with direct refinement of the detector's late task features across both seen and unseen codecs.

\subsection{Underwater Object Detection Benchmarks}
Datasets such as DUO~\cite{liu2021dataset} and RUOD~\cite{fu2023rethinking}, along with the Brackish~\cite{pedersen2019detection} and TrashCan~\cite{hong2020trashcan} datasets, provide real underwater imagery for evaluating object detection, but primarily capture degradation present at image acquisition.
They do not explicitly model the additional resolution loss and compression introduced during bandwidth-constrained transmission.
AquaTX extends DUO and RUOD with controlled resize--compression operators while preserving their original content and annotations, enabling reproducible evaluation of receiver-side restoration under both seen and unseen codec conditions.
\section{The AquaTX Benchmark}
\label{sec:benchmark}

Existing underwater object-detection benchmarks capture degradation arising from underwater image formation but generally evaluate detectors on the originally captured images.
In bandwidth-constrained deployment, however, images may be spatially reduced and compressed before reaching the receiver, introducing additional degradation that is absent from existing evaluation protocols.
This makes it difficult to isolate the effect of transmission or to compare receiver-side restoration methods under controlled conditions.
To address this gap, we introduce \textbf{AquaTX}, a benchmark that applies controlled transmission degradation to real underwater object-detection datasets while preserving their original underwater content and annotations.

\subsection{Underwater Sources and Splits}

AquaTX is constructed from two real-world underwater object-detection datasets, \textbf{DUO}~\cite{liu2021dataset} and \textbf{RUOD}~\cite{fu2023rethinking}.
Both contain imagery acquired in real underwater environments and thus naturally include degradation caused by attenuation, scattering, turbidity, and color distortion~\cite{akkaynak2018revised}.
We do not synthesize or modify these image-formation effects; AquaTX introduces only the additional digital degradation associated with transmission.

We construct two dataset-specific benchmarks, \textbf{DUO-TX} and \textbf{RUOD-TX}, using the same transmission protocol.
DUO contains four object categories (\textit{holothurian}, \textit{echinus}, \textit{scallop}, and \textit{starfish}), whereas the original RUOD contains ten categories.
For RUOD-TX, we use the four-class subset \textbf{RUOD-4C}~\cite{wille2026all}, comprising the same target categories as DUO.
This allows us to evaluate transmission-induced degradation and recovery across two underwater datasets using a consistent set of object categories.
The two datasets are trained and evaluated independently rather than merged.

For DUO, we retain the official 1,111-image test set and split its 6,671 training images into 6,003 training and 668 validation images.
For RUOD-4C, we use 1,255 test images and split its 3,013 training images into 2,712 training and 301 validation images.
Validation data are derived exclusively from the corresponding training sets, and the test sets are never used for checkpoint selection, hyperparameter tuning, or ablation decisions.
Table~\ref{tab:source_stats} summarizes the resulting dataset statistics.

\begin{table}[t]
    \centering
    \small
    \begin{tabular*}{\columnwidth}{@{\extracolsep{\fill}}llrr@{}}
    \toprule
    \textbf{Dataset} & \textbf{Split} & \textbf{Images} & \textbf{Boxes} \\
    \midrule
    \multirow{3}{*}{DUO-TX}
     & Train & 6,003 & 57,574 \\
     & Val   &   668 &  6,424 \\
     & Test  & 1,111 & 10,517 \\
    \midrule
    \multirow{3}{*}{RUOD-TX}
     & Train & 2,712 & 21,619 \\
     & Val   &   301 &  2,275 \\
     & Test  & 1,255 & 10,553 \\
    \bottomrule
    \end{tabular*}
    \caption{
    Statistics of the underwater source datasets used in AquaTX.
    DUO-TX uses the original four-class DUO dataset, while RUOD-TX uses the four-class RUOD-4C subset.
    Both are evaluated independently under the same transmission protocol.
    }
    \label{tab:source_stats}
\end{table}

\subsection{Transmission Degradation Model}

AquaTX models the digital degradation introduced when an underwater image is transmitted over a bandwidth-constrained link~\cite{ahn2018optical}.
Given a source image $x$, the received image is defined as
\begin{equation}
    x_d = T(x;L,r,\kappa),
\end{equation}
where $L$ denotes the transmitted raster size, $r$ the target coding rate, and $\kappa$ the codec.
The receiver observes only the decoded RGB image $x_d$ and is not provided with any transmission metadata.

Let $(c_w,c_h)$ denote the aspect-ratio-preserving content dimensions obtained by scaling the original image to a long side of 640 pixels.
The transmitted raster $(t_w,t_h)$ is obtained by scaling $(c_w,c_h)$ by $L/640$.
For compound degradation, each sample is generated as follows:
\begin{enumerate}
    \item \textbf{Downsample.}
    The source image is directly resized, while preserving aspect ratio, to the transmitted raster whose long side is $L$.
    \item \textbf{Encode and decode.}
    The resized image is compressed and decoded using codec $\kappa$ at target rate $r$.
    \item \textbf{Upsample.}
    The decoded raster is resized back to $(c_w,c_h)$.
    \item \textbf{Letterbox.}
    The content is centered on a $640\times640$ canvas and stored losslessly.
\end{enumerate}
Resize-only and codec-only conditions omit the compression or downscaling operation, respectively.

Codec-only controls use $L=640$, thereby isolating compression without spatial reduction.
Downsampling is performed directly from the original source using \texttt{INTER\_AREA}, rather than from an intermediate 640-pixel image.
Letterbox padding is never transmitted or codec-encoded, avoiding artificial bitrate allocation to padding.
The final decoded image is stored losslessly to prevent an additional uncontrolled coding pass.

\begin{table}[t]
\centering
\small
\setlength{\tabcolsep}{4pt}
\begin{tabular}{lccc}
\toprule
\textbf{Condition group} & \textbf{Factors} & \textbf{Validation} & \textbf{Test} \\
\midrule
Clean reference
    & - & 1 & 1 \\
Resize-only
    & 3 sizes & 3 & 3 \\
Codec-only
    & codecs$\times$3 rates & 9 & 12 \\
Compound
    & codecs$\times$3 sizes$\times$3 rates & 27 & 36 \\
\midrule
\textbf{Total}
    & & \textbf{40} & \textbf{52} \\
\bottomrule
\end{tabular}
\caption{
AquaTX evaluation conditions.
Validation uses the three seen codecs, whereas test evaluation
additionally includes the unseen AVIF codec.
}
\label{tab:conditions}
\end{table}

\subsection{Operator Family and Evaluation Conditions}

AquaTX varies three transmission factors: raster size, coding rate, and codec.

\begin{itemize}
    \item \textbf{Raster size.}
    The transmitted long side is $L\in\{256,384,512\}$, corresponding to $\{0.4,0.6,0.8\}$ of the canonical 640-pixel long side.
    Although the transmitted resolution varies, all decoded images are resized and letterboxed to the same 640×640 detector input.

    \item \textbf{Coding rate.}
    The target coding rates are $r\in\{0.30,0.60,1.00\}$ encoded bits per transmitted-raster pixel.
    Prior underwater compression studies likewise evaluate low-rate operating points along a bits-per-pixel axis~\cite{rubino2017underwater}.
    Because codec quality parameters are not directly comparable across coding standards, a per-image parameter search selects the nearest achievable rate.

    \item \textbf{Codec.}
    JPEG 4:2:0~\cite{wallace1991jpeg}, lossy WebP~\cite{rfc9649}, and irreversible JPEG2000~\cite{skodras2001jpeg} are used as
    training codecs.
    AVIF~\cite{avif_format} is excluded from training and validation and used at test time to assess unseen-codec generalization.
\end{itemize}

For auditing, AquaTX records the target rate, achieved rate, encoded size, and selected codec parameter for every generated image. 
Every evaluation image is processed under every applicable transmission condition, enabling paired comparison across degradation levels without confounding image content with transmission severity.
The evaluation protocol contains a clean reference, resize-only controls, codec-only controls, and compound resize--compression conditions, as summarized in Table~\ref{tab:conditions}.
The same protocol is applied to DUO-TX and RUOD-TX.
Reporting along an achieved-bitrate axis rather than nominal codec quality factors makes conditions comparable across coding standards. 
A fixed quality factor corresponds to different rate–distortion operating points across JPEG, WebP, and JPEG2000, so matching nominal settings would confound codec identity with degradation severity. 
The per-image parameter search instead drives each codec toward the same target bits per transmitted pixel, while the recorded achieved rate, encoded size, and selected parameter allow residual rate mismatch to be audited.
Supplementary Sec.~S3 audits this rate matching in detail, showing that the mean achieved rate remains within 0.02 bpp of the target across all codec--rate pairs.

\subsection{Frozen Evaluation and Training Protocol}

All validation and test degradations are generated once and frozen.
The same degradation-generation implementation, codec versions, and rate-matching procedure are used for DUO-TX and RUOD-TX.
We provide fixed evaluation assets with per-image transmission settings, achieved rates, and content hashes, enabling comparisons on identical decoded inputs and detection of changes to these assets.

Training degradation is generated online.
For each source dataset, training samples are drawn from a fixed mixture of 10\% clean, 10\% resize-only, 10\% codec-only, and 70\% compound conditions using only the three seen codecs.
AVIF is never used for training, validation, model selection, or hyperparameter tuning.
No transmission metadata is provided to TERA or any metadata-blind comparison method during either training or inference.


\begin{figure*}
    \centering
    \includegraphics[width=1\linewidth]{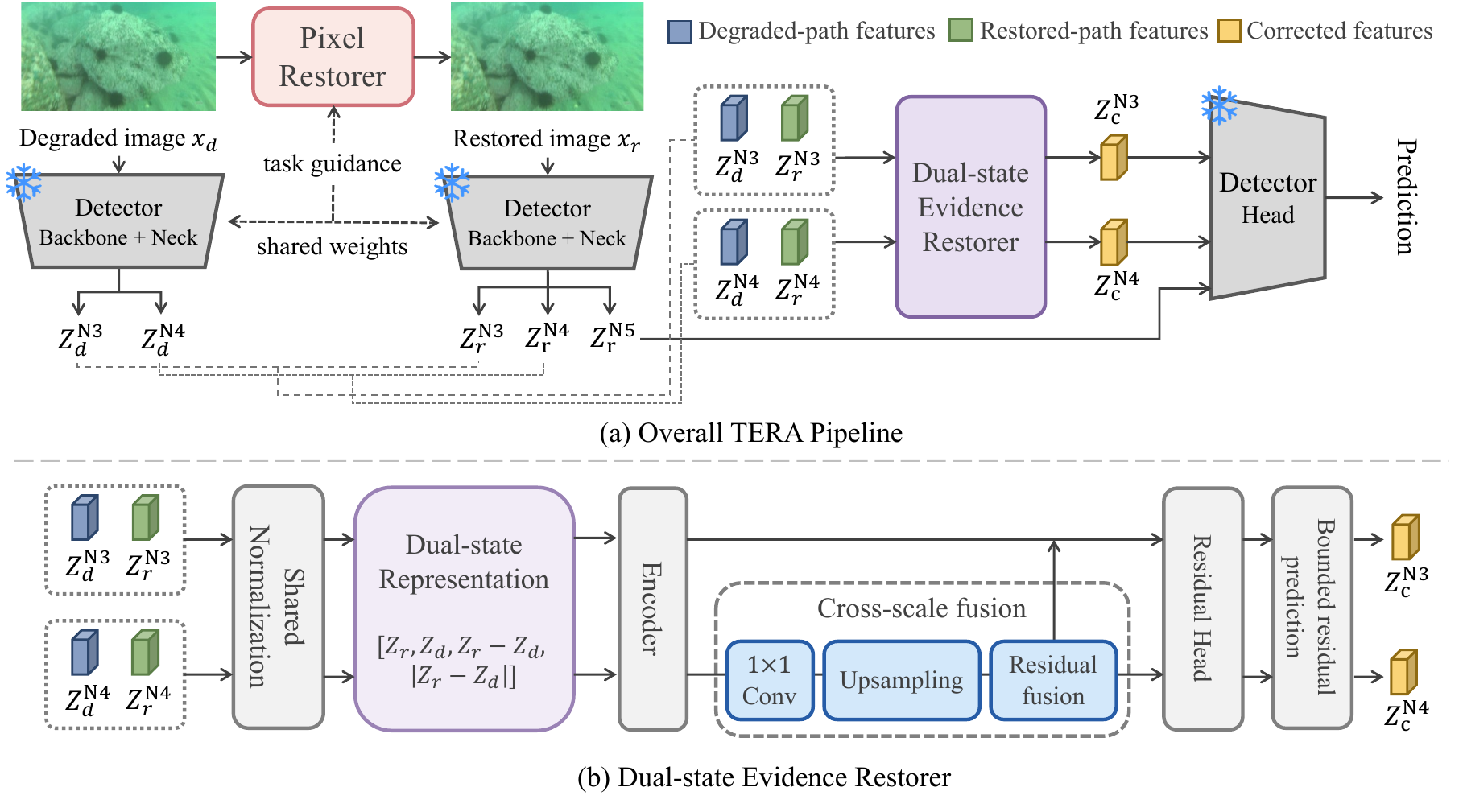}
    \caption{
    Overview of TERA.
    The pixel restorer uses intermediate backbone $P_3/P_4$ features from the degraded path for task-guided RGB restoration, whereas the evidence restorer operates on later neck features $N_3/N_4$.
    The degraded and restored images are forwarded through the same frozen backbone and neck to obtain two late feature states.
    A dual-state evidence restorer jointly exploits these states to refine the $N_3$ and $N_4$ pyramid levels, while $N_5$ follows the restored-image path unchanged.
    Only the pixel and evidence restorers are trainable.
    }
    \label{fig:tera}
\end{figure*}

\section{Method}
\label{sec:method}

In AquaTX (Sec.~\ref{sec:benchmark}), an object detector receives a transmission-degraded RGB image $x_d$ without information about the transmitted resolution, coding rate, or codec. 
The detector is trained on pre-transmission underwater images and remains frozen throughout TERA training. 
Our goal is to improve detection performance on transmission-degraded inputs without modifying the detector.

Given $x_d$, we extract $P_3/P_4$ features from the frozen detector backbone and use them to guide pixel restoration:
\begin{equation}
    E_d = \mathcal{B}_{P_3,P_4}(x_d),
    \qquad
    x_r = R_\theta(x_d,E_d),
\end{equation}
where $x_r$ is the pixel-restored image.
Here, $E_d$ denotes intermediate backbone guidance and is distinct from the late neck features used below.
Let $F^l(x)$ denote the feature at pyramid level $l$ produced by the frozen backbone and neck.
The degraded and restored paths yield
\begin{equation}
    Z_d^l = F^l(x_d),
    \quad
    Z_r^l = F^l(x_r),
    \quad
    l\in\{N_3,N_4,N_5\}.
\end{equation}

The evidence restorer jointly corrects the $N_3$ and $N_4$ levels, while $N_5$ follows the restored path unchanged:
\begin{equation}
\begin{aligned}
    \left(Z_c^{N_3},Z_c^{N_4}\right)
    &=
    E_\phi\!\left(
        Z_d^{N_3},Z_d^{N_4},
        Z_r^{N_3},Z_r^{N_4}
    \right), \\
    Z_c^{N_5} &= Z_r^{N_5}.
\end{aligned}
\end{equation}
The corrected pyramid $Z_c=\{Z_c^{N_3},Z_c^{N_4},Z_c^{N_5}\}$ is passed to the frozen detection head to obtain $\hat y$.
Only $\theta$ and $\phi$ are optimized; all detector parameters and BatchNorm statistics remain fixed.

\subsection{Pixel Restoration}

The pixel restorer $R_\theta$ is a task-guided encoder--decoder that predicts a gated RGB residual.
Its spatial gate, correction amplitude, and RGB residual are all conditioned on the degraded image and the frozen backbone guidance $E_d$:
\begin{equation}
    \begin{aligned}
        x_r = \mathrm{clip}\!\Big( x_d &+ g_s(x_d,E_d)\odot a(x_d,E_d) \\
        &\odot \Delta_{\mathrm{rgb}}(x_d,E_d), \,0,1 \Big).
    \end{aligned}
\end{equation}
The spatial gate satisfies $g_s\in[0,1]$.
We parameterize the correction amplitude as $a=a_{\max}\sigma(\cdot)$ with $a_{\max}=0.2$, and $\Delta_{\mathrm{rgb}}=\tanh(\cdot)\in[-1,1]$, bounding the per-channel RGB correction by 0.2 before clipping.
This conditioning allows detector features to guide restoration toward task-relevant image regions, while the residual formulation limits unnecessary changes to the received input.

$R_\theta$ targets spatial and coding distortions introduced during transmission rather than explicitly removing the underwater appearance already present before transmission.
Because image-space restoration may still leave discrepancies in the detector representation, TERA subsequently corrects the late task features directly.

\subsection{Dual-State Evidence Restoration}
\label{sec:method:evidence}

TERA refines the late $N_3$ and $N_4$ pyramid levels and passes the coarsest $N_5$ level unchanged from the restored path. 
The evidence restorer $E_\phi$ compares the degraded state $Z_d$ with the restored state $Z_r$ and predicts bounded residual corrections to the latter.
The restored state compensates for transmission degradation but remains imperfect, as pixel restoration may alter structures relevant to the detector.
Meanwhile, the degraded state may retain complementary evidence that is changed or suppressed by restoration.
By jointly encoding both states and their differences, $E_\phi$ conditions its correction on restoration-induced changes rather than relying on either state alone.

\paragraph{Dual-state representation with shared normalization.}
For each $l\in\{N_3,N_4\}$, level-specific per-channel statistics $(\mu^l,\sigma^l)$ are precomputed once from restored-state training features at the pixel-restoration initialization and kept fixed during joint optimization and inference. The same statistics are applied to both states:
\begin{equation}
    \tilde Z_d^l =
    \frac{Z_d^l-\mu^l}{\sigma^l},
    \qquad
    \tilde Z_r^l =
    \frac{Z_r^l-\mu^l}{\sigma^l}.
\end{equation}
Applying the same normalization to both states allows their feature differences to be measured on a consistent channel-wise scale.
Their level-wise representation is

\begin{equation}
    U^l =
    \left[
        \tilde Z_r^l,\;
        \tilde Z_d^l,\;
        \tilde Z_r^l-\tilde Z_d^l,\;
        \left|\tilde Z_r^l-\tilde Z_d^l\right|
    \right],
\end{equation}
where the signed and absolute differences encode the direction and magnitude of the change from pixel restoration.

\paragraph{Cross-scale encoding.}
Each $U^l$ is projected to 96 channels and processed by two residual blocks, producing $h_3$ and $h_4$.
A one-way $N_4\!\rightarrow\!N_3$ fusion projects and upsamples $h_4$ before combining it with $h_3$.
We denote the fused representation by $\bar{h}_{N_3}$ and set $\bar{h}_{N_4}=h_4$, without feeding the fused $N_3$ representation back into $N_4$.

\paragraph{Bounded residual prediction.}
A level-specific head $H_l$ predicts a correction in normalized feature units:
\begin{equation}
    r^l=b\,\tanh\!\left(H_l(\bar{h}_l)\right),
    \qquad l\in\{N_3,N_4\}.
\end{equation}
where $b=3$ bounds the correction magnitude.
The normalized residual is converted to the detector's native per-channel scale and added to the restored state:
\begin{equation}
    Z_c^l =
    Z_r^l+\sigma^l\odot r^l,
    \qquad l\in\{N_3,N_4\}.
\end{equation}
This bounded residual formulation corrects rather than replaces the restored detector representation.

\subsection{Training Objective}

Let $D$ denote the frozen detector and $D_{\mathrm{head}}$ its detection head.
TERA is trained with detector supervision and a pre-transmission teacher.
Let $Y_c$ and $Y_T$ denote the raw detection-head responses before box decoding and NMS for the corrected and pre-transmission reference paths, respectively.
Here, $Y^{\mathrm{cls}}$ denotes classification logits and $Y^{\mathrm{dfl}}$ the raw distributional box-regression response.
The teacher is used only during training and introduces no separately trained network.
The pre-transmission path provides a content-matched target representing the detector response before transmission-induced information loss.
It therefore guides recovery of task evidence lost during transmission rather than encouraging removal of the original underwater appearance. The overall objective is
\begin{equation}
    \mathcal{L}
    =
    \mathcal{L}_{\mathrm{GT}}
    +
    \lambda_{\mathrm{KD}}\mathcal{L}_{\mathrm{KD}}
    +
    \lambda_{\mathrm{id}}\mathcal{L}_{\mathrm{id}}
    +
    \lambda_{\mathrm{res}}\mathcal{L}_{\mathrm{res}}.
\end{equation}

\paragraph{Detection supervision.}
$\mathcal{L}_{\mathrm{GT}}$ is the detector's box-regression, classification, and distribution-focal loss~\cite{li2020generalized} computed against ground truth.
Although the detector is frozen, gradients propagate through it to the restorers.

\paragraph{Foreground-restricted teacher distillation.}
Rather than densely matching $Y_T$, we restrict distillation to the foreground set $\mathcal F$ obtained by applying the task-aligned assigner to detached corrected predictions and ground truth:
\begin{equation}
\begin{aligned}
    \mathcal{L}_{\mathrm{KD}}
    =
    \frac{1}{|\mathcal{F}|}
    \sum_{i\in\mathcal{F}}
    \Big[
        &\mathrm{BCE}\!\left(
            Y_{c,i}^{\mathrm{cls}},
            \mathrm{sg}\!\left(
                \sigma(Y_{T,i}^{\mathrm{cls}})
            \right)
        \right) \\
        &+
        \lambda_{\mathrm{dfl}}
        \mathrm{SL1}\!\left(
            Y_{c,i}^{\mathrm{dfl}},
            \mathrm{sg}(Y_{T,i}^{\mathrm{dfl}})
        \right)
    \Big],
\end{aligned}
\end{equation}
where BCE denotes binary cross-entropy with logits, SL1 denotes the smooth L1 loss, and $\mathrm{sg}(\cdot)$ denotes stop-gradient.
This focuses the teacher signal on object-related responses without forcing dense reproduction of the pre-transmission detector output.
This avoids allowing the large number of background locations to dominate the distillation objective.

\paragraph{Identity and residual regularization.}
To discourage unnecessary feature correction when no transmission degradation is present, we regularize the residual $r_{\mathrm{ref}}^l$ predicted for transmission-free inputs:
\begin{equation}
    \mathcal{L}_{\mathrm{id}}
    =
    \sum_{l\in\{N_3,N_4\}}
    \left\|
        r_{\mathrm{ref}}^l
    \right\|_1.
\end{equation}
We additionally regularize the magnitude of feature corrections on degraded inputs:
\begin{equation}
    \mathcal{L}_{\mathrm{res}}
    =
    \sum_{l\in\{N_3,N_4\}}
    \left\|
        r^l
    \right\|_1.
\end{equation}
These terms bias the evidence restorer toward small corrections while allowing the detector-driven objectives to determine their task-relevant direction.
The pixel residual is not explicitly regularized; its magnitude is controlled structurally by the bounded amplitude term, spatial gating, and zero-initialized residual head.

The pixel and evidence restorers are optimized jointly while the detector remains frozen.
At inference time, the teacher and all training losses are removed;
TERA requires only the received degraded image and the frozen detector.
\section{Experiments}
\label{sec:experiments}

\begin{table*}[t]
    \centering
    \small
    \setlength{\tabcolsep}{3pt}
    \begin{tabularx}{\textwidth}{
        @{}>{\raggedright\arraybackslash}p{0.24\textwidth}
        >{\raggedright\arraybackslash}X cccccc
    }
    \toprule
    \textbf{Strategy} & \textbf{Method}
    & \textbf{Clean}
    & \textbf{Resize-only}
    & \textbf{Codec-only}
    & \textbf{Seen comp.}
    & \textbf{Unseen AVIF}
    & \textbf{$\Delta$Clean} \\
    & & (1) & (3) & (9) & (27) & (9) & \\
    \midrule

    \multicolumn{8}{@{}l}{\textbf{DUO-TX}} \\
    \midrule

    No adaptation & Clean-YOLO
    & \underline{65.7} & 60.3 & 57.0 & 37.7 & 42.2 & \underline{0.0} \\

    \addlinespace[2pt]

    \multirow{3}{*}{Image restoration}
    & UWCNN~\cite{li2020underwater} (FT)
    & 63.4 & 59.6 & 56.0 & 38.3 & 43.0 & -2.3 \\

    & NU2Net~\cite{guo2023underwater} (FT)
    & 64.2 & \underline{60.4} & 56.5 & 38.9 & 43.4 & -1.5 \\

    & EDTR (FT)~\cite{kim2025exploiting}
    & 33.6 & 33.2 & 32.4 & 27.3 & 28.7 & -32.1 \\

    \addlinespace[2pt]

    Detector adaptation
    & DegFT-YOLO
    & 61.4 & 59.9 & \underline{58.2}
    & \underline{48.1} & \underline{50.8} & -4.3 \\

    \addlinespace[2pt]
    \rowcolor{gray!10}
    Task-evidence restoration
    & \textbf{TERA (Ours)}
    & \textbf{66.4} & \textbf{65.7} & \textbf{61.9}
    & \textbf{50.0} & \textbf{53.2} & \textbf{+0.7} \\

    \midrule

    \multicolumn{8}{@{}l}{\textbf{RUOD-TX}} \\
    \midrule

    No adaptation & Clean-YOLO
    & \underline{51.5} & 48.4 & 46.0 & 31.5 & 35.7 & \underline{0.0} \\

    \addlinespace[2pt]

    \multirow{3}{*}{Image restoration}
    & UWCNN (FT)
    & 40.8 & 37.6 & 36.2 & 24.9 & 28.1 & -10.7 \\

    & NU2Net (FT)
    & 50.6 & \underline{48.7} & 45.8 & 32.4 & 36.1 & -0.9 \\

    & EDTR (FT)
    & 24.1 & 23.9 & 23.7 & 20.1 & 20.8 & -27.4 \\

    \addlinespace[2pt]

    Detector adaptation
    & DegFT-YOLO
    & 48.6 & 48.0 & \underline{47.0}
    & \textbf{41.4} & \underline{43.1} & -2.9 \\

    \addlinespace[2pt]
    \rowcolor{gray!10}
    Task-evidence restoration
    & \textbf{TERA (Ours)}
    & \textbf{52.1} & \textbf{51.0} & \textbf{49.2}
    & \underline{41.3} & \textbf{43.2} & \textbf{+0.6} \\

    \bottomrule
    \end{tabularx}

    \caption{
        Main detection results on AquaTX in mAP$_{50:95}$ (\%).
        Clean-YOLO is trained on the original pre-transmission images and directly evaluated without restoration or degradation adaptation.
        Image-restoration methods operate in front of the same frozen Clean-YOLO detector, whereas DegFT-YOLO adapts the detector itself.
        Resize-only, codec-only, and seen-compound results are macro-averaged over 3, 9, and 27 conditions, respectively.
        Unseen AVIF denotes the nine compound conditions generated using the held-out AVIF codec; the three AVIF codec-only controls are excluded from the main aggregate.
        $\Delta$Clean is the clean-input change relative to Clean-YOLO.
        Best and second-best results are shown in bold and underline, respectively.
        Zero-shot results and additional enhancement baselines are reported in the supplementary material.
    }
    \label{tab:main}
\end{table*}

\subsection{Implementation Details}

\paragraph{Detector and baselines.}
All experiments use YOLOv8s~\cite{yolov8_ultralytics} (11.1M parameters) at $640\times640$ resolution.
We train \emph{Clean-YOLO} on the pre-transmission images for 100 epochs and keep its weights and BatchNorm statistics frozen for TERA and receiver-side restoration baselines.
The degradation-fine-tuned baseline (\emph{DegFT-YOLO}) is initialized from Clean-YOLO and fine-tuned for 50 epochs using the same AquaTX training degradation distribution as TERA.
Clean-YOLO uses mosaic, geometric (scaling and translation), color (HSV), and horizontal-flip augmentations, whereas DegFT-YOLO uses only horizontal flipping.
We report mAP$_{50:95}$, averaged over IoU thresholds from 0.50 to 0.95 in steps of 0.05.

\paragraph{TERA training.}
TERA contains a 6.52M-parameter pixel restorer based on a NAFNet encoder--decoder~\cite{chen2022simple} and a 1.05M-parameter evidence restorer.
The evidence restorer operates on $N_3/N_4$ with width-96 dual-state encoders and one-way $N_4\!\rightarrow\!N_3$ fusion, while $N_5$ remains unchanged.
Joint TERA optimization is initialized from a pretrained pixel-restoration checkpoint.
The pixel and evidence restorers are then optimized together with AdamW~\cite{loshchilov2017decoupled} and cosine annealing~\cite{loshchilov2016sgdr} for 6,000 steps on DUO-TX and 5,000 steps on RUOD-TX.
We set $\lambda_{\mathrm{KD}}=1.0$, $\lambda_{\mathrm{dfl}}=0.1$, $\lambda_{\mathrm{res}}=10^{-3}$, and $\lambda_{\mathrm{id}}=0.5$.
Additional training details are provided in the supplementary material.

\paragraph{Computational profiling.}
Full TERA uses two frozen backbone/neck passes for the degraded and restored inputs and runs at 25.81 ms/image for $640\times640$ inputs with batch size 1 on an NVIDIA H200.
For comparison, the frozen detector alone requires 4.99 ms/image, while pixel-only restoration takes 23.28 ms/image, indicating that the additional dual-state feature stage incurs approximately 2.5 ms.
This cost reflects a receiver-side compute--robustness trade-off and does not increase the transmitted data volume.
Detailed comparisons of parameter count, FLOPs, latency, peak GPU memory, and the accuracy--compute trade-off are provided in the supplementary material.

\subsection{Main Comparison}
\label{sec:results:main}

Transmission degradation substantially reduces detection performance on both datasets.
Under the primary compound setting, mAP$_{50:95}$ drops from 65.7 to 37.7 on DUO-TX and from 51.5 to 31.5 on RUOD-TX.

\paragraph{Comparison with detector adaptation.}
DegFT recovers substantial robustness but reduces clean-input performance from 65.7 to 61.4 on DUO-TX and from 51.5 to 48.6 on RUOD-TX.
TERA instead keeps the detector frozen, outperforming DegFT by 1.9 points on DUO-TX and achieving essentially identical compound performance on RUOD-TX.

Under the isolated transmission factors, TERA outperforms DegFT under both resize-only and codec-only degradation on DUO-TX (65.7/61.9 vs.\ 59.9/58.2) and RUOD-TX (51.0/49.2 vs.\ 48.0/47.0).

\begin{figure}
    \centering
    \includegraphics[width=1\linewidth]{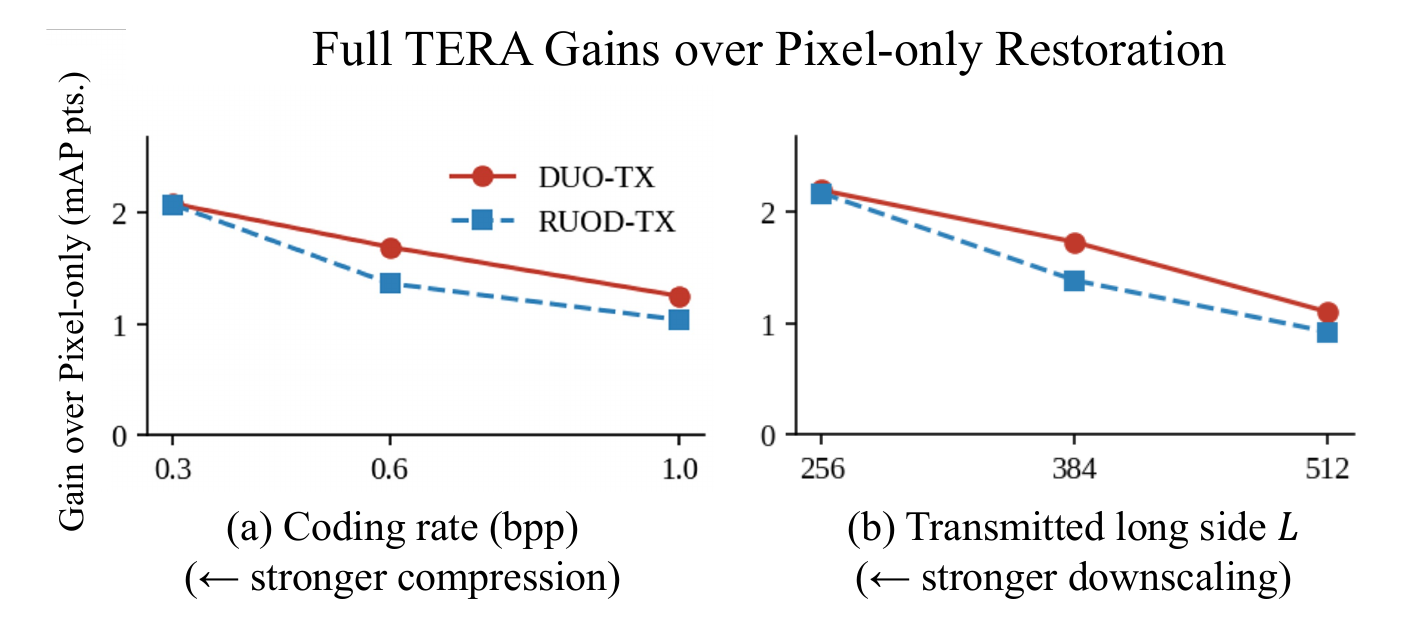}
    \caption{
    Incremental gain of full TERA over pixel-only restoration under varying transmission severity.
    When averaged over the remaining transmission factors, the gain over pixel-only restoration is larger at lower coding rates (left) and smaller transmitted rasters (right) on both datasets.
    }
    \label{fig:severity}
\end{figure}

\paragraph{Comparison with receiver-side restoration.}
Even after task-specific fine-tuning, conventional receiver-side restoration baselines remain limited under compound transmission.
For example, NU2Net-FT reaches 38.9 and 32.4 mAP$_{50:95}$ on DUO-TX and RUOD-TX, respectively, compared with 50.0 and 41.3 for TERA.
Zero-shot application of existing enhancement and restoration methods consistently performs below the raw transmitted input; detailed results are provided in the supplementary material.

\begin{figure*}
    \centering
    \includegraphics[width=1\linewidth]{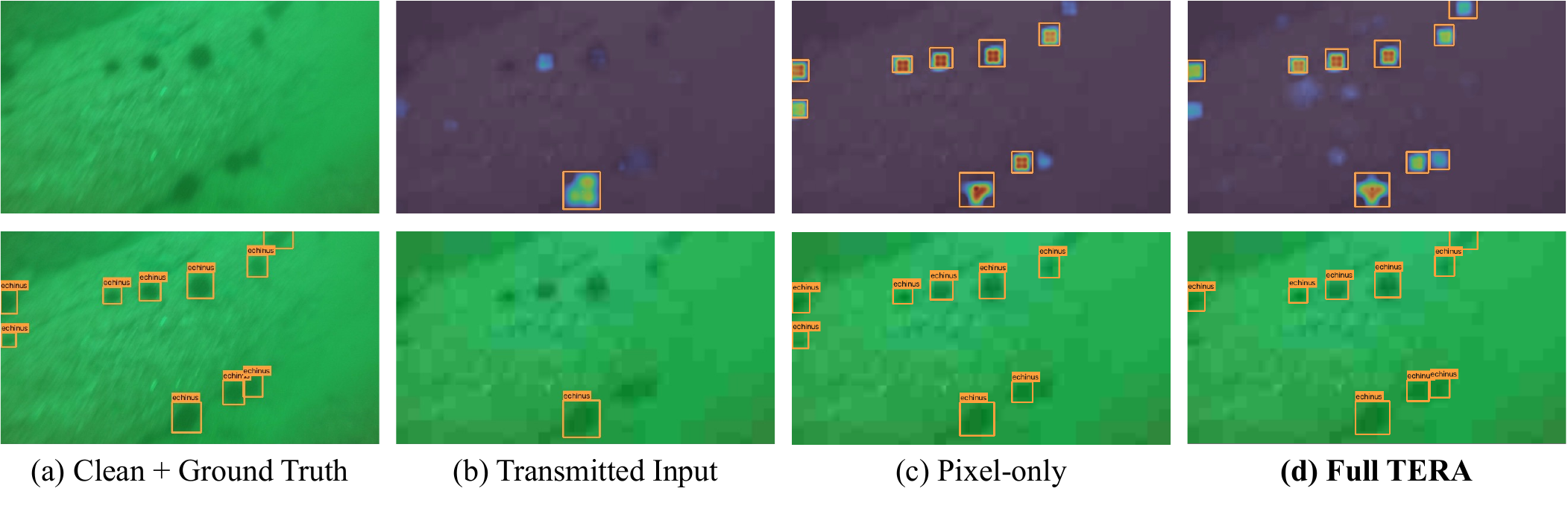}
    \caption{
        Qualitative ablation of TERA.
        Columns (a)--(d) show the clean reference with ground truth, the transmitted input, the result after pixel-only restoration, and the final detection result obtained using full TERA, respectively.
        The first row visualizes detector response maps before box decoding, where warmer colors indicate stronger object evidence.
        The second row shows the corresponding detection results.
        Compared with pixel-only restoration, full TERA strengthens object responses and recovers additional detections.
    }
    \label{fig:placeholder}
\end{figure*}

\paragraph{Generalization to an unseen codec.}
TERA achieves 53.2 on DUO-TX and 43.2 on RUOD-TX under the unseen AVIF codec, compared with 50.8 and 43.1 for DegFT.
Since DegFT also generalizes, TERA demonstrates comparable or better robustness under codec shift.

\subsection{Ablation Study}
\label{sec:ablation}

We ablate the main components of TERA on DUO-TX in Table~\ref{tab:ablation}.
Pixel restoration provides the largest single improvement, increasing seen-compound mAP$_{50:95}$ from 37.67 to 48.33 and unseen-codec performance from 42.19 to 51.58.
This establishes image-space restoration as the primary stage while maintaining clean-input performance.

\begin{table}[t]
    \centering
    \small
    \setlength{\tabcolsep}{3.5pt}
    \begin{tabular*}{\linewidth}{@{\extracolsep{\fill}}lccc@{}}
    \toprule
    \textbf{Configuration} & \textbf{Seen} & \textbf{Clean} & \textbf{Unseen} \\
    & \textbf{Comp.} & & \textbf{AVIF} \\
    \midrule
    Raw & 37.7 & 65.7 & 42.2 \\
    Pixel only & 48.3 & 66.0 & 51.6 \\
    + Dual-state restoration & 49.6 & \textbf{66.6} & 52.8 \\
    \textbf{+ Foreground KD (TERA)} & \textbf{50.0} & 66.4 & \textbf{53.2} \\
    Feature-L1 objective & 45.9 & 65.5 & 49.9 \\
    \bottomrule
    \end{tabular*}
    \caption{
    Ablation of TERA on DUO-TX in mAP$_{50:95}$ (\%).
    Seen Comp. and Unseen AVIF are macro-averaged over 27 and 9 conditions, respectively.
    }
    \label{tab:ablation}
\end{table}

Adding dual-state evidence restoration further improves the two settings to 49.58 and 52.81, respectively, indicating that detector-space correction recovers residual task evidence not recovered by pixel restoration alone.
Figure~\ref{fig:severity} further shows that, when averaged by transmission factor, the gain of full TERA over pixel-only restoration increases with transmission severity.
The average gain over pixel-only restoration is larger at lower coding rates and smaller transmitted rasters on both datasets, reaching 2.07 and 2.17 mAP points on RUOD-TX at 0.30 bpp and $L=256$, respectively.
Notably, full TERA outperforms pixel-only restoration in all 27 seen-compound conditions on both datasets, supporting the effectiveness of the complete refinement approach under diverse transmission conditions.
This improvement comes at an additional inference cost of 2.53 ms/image over pixel-only restoration (25.81 vs. 23.28 ms on an H200).
Foreground-restricted teacher distillation provides a smaller additional gain, improving seen-compound performance from 49.58 to 50.0.
Because the teacher is used only during training, this gain introduces no additional inference-time computation.

We further compare the task-driven objective with direct L1 reconstruction of the pre-transmission detector features.
Feature-L1 matching reaches only 45.94 under seen compound degradation and 49.86 on the unseen codec, performing below pixel restoration alone.
Thus, minimizing a generic distance to the pre-transmission representation is not sufficient for recovering features useful to the detection head, supporting task-driven evidence restoration.
\section{Conclusion}
\label{sec:conclusion}

We introduced AquaTX and TERA to study underwater object detection under bandwidth-constrained transmission.
AquaTX enables controlled evaluation of transmission-induced degradation through independent underwater datasets, factor-isolated controls, and a held-out codec.
TERA restores task evidence without modifying the detector or requiring transmission metadata.
It uses frozen backbone guidance for pixel restoration and combines degraded and restored detector features to recover residual evidence in the late feature pyramid.
Across DUO-TX and RUOD-TX, TERA substantially recovers detection performance while preserving clean-input performance and remaining competitive under a held-out codec.
Unlike detector fine-tuning, TERA does not require sacrificing clean-input accuracy.
Ablation results show image-space restoration provides the primary recovery, while feature refinement and foreground distillation provide additional gains, particularly under stronger transmission degradation.
These results highlight the importance of restoring task-relevant representations rather than image appearance alone.

This study is limited to a single detector architecture, leaving its generalizability to other detector families unverified.
Moreover, AquaTX does not model packet loss or time-varying channel conditions.
Additionally, TERA introduces inference overhead that may limit deployment on resource-constrained hardware.
Future work will explore efficient restoration and broader evaluation across detectors, codecs, and transmission conditions.

\section*{Acknowledgements}
This work was supported by the National Research Foundation of Korea (NRF) grant funded by the Korea government (MSIT) (RS-2026-25617165). This work also was supported by the Institute of Information \& Communications Technology Planning \& Evaluation (IITP) grant funded by the Korea government (MSIT) (RS-2026-25616928, AI Star Fellowship Support Program (Ulsan National Institute of Science and Technology, Inter-University); RS-2024-00436936, the ITRC (Information Technology Research Center) grant; RS-2023-00259497, the ICAN (ICT Challenge and Advanced Network of HRD) support program).

{
    \small
    \bibliographystyle{ieeenat_fullname}
    \bibliography{main}

@String(CVPR= {IEEE Conf. Comput. Vis. Pattern Recog.})

@String(ICCV= {Int. Conf. Comput. Vis.})

@String(ECCV= {Eur. Conf. Comput. Vis.})

@String(ICPR = {Int. Conf. Pattern Recog.})

@String(ICIP = {IEEE Int. Conf. Image Process.})

@String(AAAI = {AAAI})

@String(CVPR  = {CVPR})

@String(ICCV  = {ICCV})

@String(ECCV  = {ECCV})

@String(ICPR  = {ICPR})

@String(ICIP  = {ICIP})

@inproceedings{liu2021dataset,
  title={A dataset and benchmark of underwater object detection for robot picking},
  author={Liu, Chongwei and Li, Haojie and Wang, Shuchang and Zhu, Ming and Wang, Dong and Fan, Xin and Wang, Zhihui},
  booktitle={Proceedings of the IEEE International Conference on Multimedia \& Expo Workshops (ICMEW)},
  pages={1--6},
  year={2021},
  organization={IEEE}
}

@article{fu2023rethinking,
  title={Rethinking general underwater object detection: Datasets, challenges, and solutions},
  author={Fu, Chenping and Liu, Risheng and Fan, Xin and Chen, Puyang and Fu, Hao and Yuan, Wanqi and Zhu, Ming and Luo, Zhongxuan},
  journal={Neurocomputing},
  volume={517},
  pages={243--256},
  year={2023},
  publisher={Elsevier}
}

@software{yolov8_ultralytics,
  author = {Glenn Jocher and Ayush Chaurasia and Jing Qiu},
  title = {Ultralytics {YOLOv8}},
  version = {8.0.0},
  year = {2023},
  url = {https://github.com/ultralytics/ultralytics},
  license = {AGPL-3.0}
}

@article{wallace1991jpeg,
  title={The JPEG still picture compression standard},
  author={Wallace, Gregory K},
  journal={Communications of the ACM},
  volume={34},
  number={4},
  pages={30--44},
  year={1991}
}

@article{skodras2001jpeg,
  title={The JPEG 2000 still image compression standard},
  author={Skodras, Athanassios and Christopoulos, Charilaos and Ebrahimi, Touradj},
  journal={IEEE Signal processing magazine},
  volume={18},
  number={5},
  pages={36--58},
  year={2001},
  publisher={IEEE}
}

@article{liu2022twin,
  title={Twin adversarial contrastive learning for underwater image enhancement and beyond},
  author={Liu, Risheng and Jiang, Zhiying and Yang, Shuzhou and Fan, Xin},
  journal={IEEE Transactions on Image Processing},
  volume={31},
  pages={4922--4936},
  year={2022},
  publisher={IEEE}
}

@article{lin2026downstream,
  title={Downstream task inspired underwater image enhancement: a perception-aware study from dataset construction to network design},
  author={Lin, Bosen and Gao, Feng and Yu, Yanwei and Dong, Junyu and Du, Qian},
  journal={IEEE Transactions on Image Processing},
  year={2026},
  publisher={IEEE},
  volume={35},
    pages={2873--2888}
}

@article{er2023research,
  title={Research challenges, recent advances, and popular datasets in deep learning-based underwater marine object detection: A review},
  author={Er, Meng Joo and Chen, Jie and Zhang, Yani and Gao, Wenxiao},
  journal={Sensors},
  volume={23},
  number={4},
  pages={1990}
}

@article{liu2021new,
  title={A new dataset, {Poisson GAN} and {AquaNet} for underwater object grabbing},
  author={Liu, Chongwei and Wang, Zhihui and Wang, Shijie and Tang, Tao and Tao, Yulong and Yang, Caifei and Li, Haojie and Liu, Xing and Fan, Xin},
  journal={IEEE Transactions on Circuits and Systems for Video Technology},
  volume={32},
  number={5},
  pages={2831--2844},
  year={2021}
}

@article{ditria2020automating,
  title={Automating the analysis of fish abundance using object detection: optimizing animal ecology with deep learning},
  author={Ditria, Ellen M and Lopez-Marcano, Sebastian and Sievers, Michael and Jinks, Eric L and Brown, Christopher J and Connolly, Rod M},
  journal={Frontiers in Marine Science},
  volume={7},
  pages={429},
  year={2020}
}

@article{wu2025underwater,
  title={Underwater image compression for human and machine visions with hybrid priors embedding},
  author={Wu, Jianhao and Lu, Zhanhong and Mao, Yudong and Wang, Shanshan and Jiang, Qiuping},
  journal={Pattern Recognition},
  volume = {172},
  pages={112754},
  year={2026},
  publisher={Elsevier}
}

@article{yuan2021low,
  title={Low bit-rate compression of underwater image based on human visual system},
  author={Yuan, Fei and Zhan, Lihui and Pan, Panwang and Cheng, En},
  journal={Signal Processing: Image Communication},
  volume={91},
  pages={116082},
  year={2021},
  publisher={Elsevier}
}

@inproceedings{li2023rfd,
    author    = {Li, Mengyao and Shen, Liquan and Ye, Peng and Feng, Guorui and Wang, Zheyin},
    title     = {{RFD-ECNet}: Extreme Underwater Image Compression with Reference to Feature Dictionary},
    booktitle = {Proceedings of the IEEE/CVF International Conference on Computer Vision (ICCV)},
    month     = {October},
    year      = {2023},
    pages     = {12980--12989}
    }

@article{zhu2026toward,
  title={Toward Efficient Underwater Visual Perception through Image Enhancement, Compression, and Understanding},
  author={Zhu, Rongxin and Sheng, Lei and Wu, Kaitao and Boukerche, Azzedine and Long, Libo and Yang, Qiuling},
  journal={ACM Computing Surveys},
  volume={58},
  number={8},
  pages={1--46},
  year={2026},
  publisher={ACM New York, NY}
}

@article{gandor2022first,
  title={First gradually, then suddenly: understanding the impact of image compression on object detection using deep learning},
  author={Gandor, Tomasz and Nalepa, Jakub},
  journal={Sensors},
  volume={22},
  number={3},
  pages={1104},
  year={2022},
  publisher={MDPI}
}

@article{zhang2024deep,
  title={Deep underwater image compression for enhanced machine vision applications},
  author={Zhang, Hanshu and Fan, Suzhen and Zou, Shuo and Yu, Zhibin and Zheng, Bing},
  journal={Frontiers in Marine Science},
  volume={11},
  pages={1411527},
  year={2024},
  publisher={Frontiers Media SA}
}

@article{zheng2026comprehensive,
  title = {A comprehensive review of underwater image enhancement methods},
    journal = {Neural Networks},
    volume = {205},
    pages = {109527},
    year = {2027},
    issn = {0893-6080},
    doi = {https://doi.org/10.1016/j.neunet.2026.109527},
    url = {https://www.sciencedirect.com/science/article/pii/S0893608026009858},
    author = {Ying Zheng and Yifan Feng and Ran Liu and Miao Yu and Songlin Jin and Zheng Liang and Deguang Li and Weidong Zhang},
}

@misc{rfc9649,
    series =    {Request for Comments},
    number =    9649,
    howpublished =  {RFC 9649},
    publisher = {RFC Editor},
    doi =       {10.17487/RFC9649},
    url =       {https://www.rfc-editor.org/info/rfc9649/},
    author =    {J. Zern and P. Massimino and J. Alakuijala},
    title =     {{WebP Image Format}},
    pagetotal = 46,
    year =      2024,
    month =     nov,
}

@article{chen2020perceptual,
  title={Perceptual underwater image enhancement with deep learning and physical priors},
  author={Chen, Long and Jiang, Zheheng and Tong, Lei and Liu, Zhihua and Zhao, Aite and Zhang, Qianni and Dong, Junyu and Zhou, Huiyu},
  journal={IEEE Transactions on Circuits and Systems for Video Technology},
  volume={31},
  number={8},
  pages={3078--3092},
  year={2020},
  publisher={IEEE}
}

@inproceedings{kim2025exploiting,
    author    = {Kim, Jaeha and Oh, Junghun and Lee, Kyoung Mu},
    title     = {Exploiting Diffusion Prior for Task-driven Image Restoration},
    booktitle = {Proceedings of the IEEE/CVF International Conference on Computer Vision (ICCV)},
    month     = {October},
    year      = {2025},
    pages     = {10151--10161}
    }

@inproceedings{haris2021task,
  title={Task-driven super resolution: Object detection in low-resolution images},
  author={Haris, Muhammad and Shakhnarovich, Greg and Ukita, Norimichi},
  booktitle={International conference on neural information processing},
  pages={387--395},
  year={2021},
  organization={Springer}
}

@inproceedings{xie2019feature,
  title={Feature denoising for improving adversarial robustness},
  author={Xie, Cihang and Wu, Yuxin and Maaten, Laurens van der and Yuille, Alan L and He, Kaiming},
  booktitle={Proceedings of the IEEE/CVF conference on computer vision and pattern recognition},
  pages={501--509},
  year={2019}
}

@inproceedings{li2026hpgn,
  title={{HPGN}: Hybrid Priors-Guided Network for Compressed Low-Light Image Enhancement},
  author={Li, Hantang and Zhu, Qiang and Meng, Xiandong and Xiong, Lei and Zhu, Shuyuan and Fan, Xiaopeng},
  booktitle={2026 IEEE International Conference on Image Processing (ICIP)},
  pages={1--6},
  year={2026},
  organization={IEEE}
}

@article{ahn2018optical,
  title={An optical image transmission system for deep sea creature sampling missions using autonomous underwater vehicle},
  author={Ahn, Jonghyun and Yasukawa, Shinsuke and Sonoda, Takashi and Nishida, Yuya and Ishii, Kazuo and Ura, Tamaki},
  journal={IEEE Journal of Oceanic Engineering},
  volume={45},
  number={2},
  pages={350--361},
  year={2018},
  publisher={IEEE}
}

@article{rubino2017underwater,
  title={Underwater radio frequency image sensor using progressive image compression and region of interest},
  author={Rubino, Eduardo M and Centelles, Diego and Sales, Jorge and Marti, Jose V and Marin, Raul and Sanz, Pedro J and Alvares, Alberto J},
  journal={Journal of the Brazilian Society of Mechanical Sciences and Engineering},
  volume={39},
  number={10},
  pages={4115--4134},
  year={2017},
  publisher={Springer}
}

@article{li2020underwater,
  title={Underwater scene prior inspired deep underwater image and video enhancement},
  author={Li, Chongyi and Anwar, Saeed and Porikli, Fatih},
  journal={Pattern recognition},
  volume={98},
  pages={107038},
  year={2020},
  publisher={Elsevier}
}

@inproceedings{guo2023underwater,
  title={Underwater ranker: Learn which is better and how to be better},
  author={Guo, Chunle and Wu, Ruiqi and Jin, Xin and Han, Linghao and Zhang, Weidong and Chai, Zhi and Li, Chongyi},
  booktitle={Proceedings of the AAAI conference on artificial intelligence},
  volume={37},
  number={1},
  pages={702--709},
  year={2023}
}

@article{li2020generalized,
  title={Generalized focal loss: Learning qualified and distributed bounding boxes for dense object detection},
  author={Li, Xiang and Wang, Wenhai and Wu, Lijun and Chen, Shuo and Hu, Xiaolin and Li, Jun and Tang, Jinhui and Yang, Jian},
  journal={Advances in neural information processing systems},
  volume={33},
  pages={21002--21012},
  year={2020}
}

@article{li2019underwater,
  title={An underwater image enhancement benchmark dataset and beyond},
  author={Li, Chongyi and Guo, Chunle and Ren, Wenqi and Cong, Runmin and Hou, Junhui and Kwong, Sam and Tao, Dacheng},
  journal={IEEE transactions on image processing},
  volume={29},
  pages={4376--4389},
  year={2019},
  publisher={IEEE}
}

@article{li2021underwater,
  title={Underwater image enhancement via medium transmission-guided multi-color space embedding},
  author={Li, Chongyi and Anwar, Saeed and Hou, Junhui and Cong, Runmin and Guo, Chunle and Ren, Wenqi},
  journal={IEEE Transactions on Image Processing},
  volume={30},
  pages={4985--5000},
  year={2021},
  publisher={IEEE}
}

@article{peng2023u,
  title={U-shape transformer for underwater image enhancement},
  author={Peng, Lintao and Zhu, Chunli and Bian, Liheng},
  journal={IEEE transactions on image processing},
  volume={32},
  pages={3066--3079},
  year={2023},
  publisher={IEEE}
}

@inproceedings{loshchilov2017decoupled,
    title={Decoupled Weight Decay Regularization},
    author={Ilya Loshchilov and Frank Hutter},
    booktitle={International Conference on Learning Representations},
    year={2019},
    url={https://openreview.net/forum?id=Bkg6RiCqY7},
}

@inproceedings{loshchilov2016sgdr,
    title={{SGDR}: Stochastic Gradient Descent with Warm Restarts},
    author={Ilya Loshchilov and Frank Hutter},
    booktitle={International Conference on Learning Representations},
    year={2017},
    url={https://openreview.net/forum?id=Skq89Scxx}
}

@inproceedings{akkaynak2019sea,
  title={Sea-Thru: A Method for Removing Water From Underwater Images},
  author={Akkaynak, Derya and Treibitz, Tali},
  booktitle={Proceedings of 2019 IEEE/CVF Conference on Computer Vision and Pattern Recognition (CVPR)},
  pages={1682--1691},
  year={2019},
  organization={IEEE}
}

@article{xu2023systematic,
  title={A systematic review and analysis of deep learning-based underwater object detection},
  author={Xu, Shubo and Zhang, Minghua and Song, Wei and Mei, Haibin and He, Qi and Liotta, Antonio},
  journal={Neurocomputing},
  volume={527},
  pages={204--232},
  year={2023},
  publisher={Elsevier}
}

@article{jian2024underwater,
  title={Underwater object detection and datasets: a survey},
  author={Jian, Muwei and Yang, Nan and Tao, Chen and Zhi, Huixiang and Luo, Hanjiang},
  journal={Intelligent Marine Technology and Systems},
  volume={2},
  number={1},
  pages={9},
  year={2024},
  publisher={Springer}
}

@inproceedings{dodge2016understanding,
  title={Understanding how image quality affects deep neural networks},
  author={Dodge, Samuel and Karam, Lina},
  booktitle={2016 eighth international conference on quality of multimedia experience (QoMEX)},
  pages={1--6},
  year={2016},
  organization={IEEE}
}

@inproceedings{poyser2021impact,
  title={On the impact of lossy image and video compression on the performance of deep convolutional neural network architectures},
  author={Poyser, Matt and Atapour-Abarghouei, Amir and Breckon, Toby P},
  booktitle={2020 25th international conference on pattern recognition (ICPR)},
  pages={2830--2837},
  year={2021},
  organization={IEEE}
}

@article{anwar2020diving,
  title={Diving deeper into underwater image enhancement: A survey},
  author={Anwar, Saeed and Li, Chongyi},
  journal={Signal Processing: Image Communication},
  volume={89},
  pages={115978},
  year={2020},
  publisher={Elsevier}
}

@inproceedings{yang2022self,
  title={Self-feature distillation with uncertainty modeling for degraded image recognition},
  author={Yang, Zhou and Dong, Weisheng and Li, Xin and Wu, Jinjian and Li, Leida and Shi, Guangming},
  booktitle = {Computer Vision – ECCV 2022: 17th European Conference, Tel Aviv, Israel, October 23–27, 2022, Proceedings, Part XXIV},
  pages={552--569},
  year={2022},
  organization={Springer}
}

@InProceedings{pedersen2019detection,
    author = {Pedersen, Malte and Bruslund Haurum, Joakim and Gade, Rikke and Moeslund, Thomas B.},
    title = {Detection of Marine Animals in a New Underwater Dataset with Varying Visibility},
    booktitle = {Proceedings of the IEEE/CVF Conference on Computer Vision and Pattern Recognition (CVPR) Workshops},
    pages={18--26},
    month = {June},
    year = {2019}
}

@misc{hong2020trashcan,
  author = {Fulton, Michael S. and Hong, Jungseok and Sattar, Junaed},
  title = {{TrashCan 1.0}: An Instance-Segmentation Labled Dataset of Trash Observations},
  year = {2020},
  publisher = {Data Repository for the University of Minnesota},
  doi = {10.13020/g1gx-y834},
  url = {https://doi.org/10.13020/g1gx-y834},
  howpublished={Data Repository for the University of Minnesota},
  note={doi:10.13020/g1gx-y834}
}

@article{akyildiz2005underwater,
  title={Underwater acoustic sensor networks: research challenges},
  author={Akyildiz, Ian F and Pompili, Dario and Melodia, Tommaso},
  journal={Ad hoc networks},
  volume={3},
  number={3},
  pages={257--279},
  year={2005},
  publisher={Elsevier}
}

@inproceedings{liu2022image,
  title={Image-adaptive YOLO for object detection in adverse weather conditions},
  author={Liu, Wenyu and Ren, Gaofeng and Yu, Runsheng and Guo, Shi and Zhu, Jianke and Zhang, Lei},
  booktitle={Proceedings of the AAAI conference on artificial intelligence},
  volume={36},
  number={2},
  pages={1792--1800},
  year={2022}
}

@misc{avif_format,
  author       = {{Alliance for Open Media}},
  title        = {{AV1} Image File Format ({AVIF})},
  year         = {2019},
  howpublished = {\url{https://aomediacodec.github.io/av1-avif/}},
  note         = {Accessed: 2026-08-27}
}

@inproceedings{akkaynak2018revised,
    author = {Akkaynak, Derya and Treibitz, Tali},
    title = {A Revised Underwater Image Formation Model},
    booktitle = {Proceedings of the IEEE Conference on Computer Vision and Pattern Recognition (CVPR)},
    month = {June},
    year = {2018}
    }

@inproceedings{wille2026all,
    author    = {Wille, Melanie and Fischer, Tobias and Raine, Scarlett},
    title     = {Are All Marine Species Created Equal? Performance Disparities in Underwater Object Detection},
    booktitle = {Proceedings of the IEEE/CVF Winter Conference on Applications of Computer Vision (WACV)},
    month     = {March},
    year      = {2026},
    pages     = {4556--4565}
    }

@inproceedings{chen2022simple,
  title={Simple baselines for image restoration},
  author={Chen, Liangyu and Chu, Xiaojie and Zhang, Xiangyu and Sun, Jian},
  booktitle = {Computer Vision – ECCV 2022: 17th European Conference, Tel Aviv, Israel, October 23–27, 2022, Proceedings, Part VII},
  pages={17--33},
  year={2022},
  organization={Springer}
}
}

\end{document}